\pdfoutput=1

\documentclass[11pt]{article}

\usepackage{naacl2021}

\usepackage{times}
\usepackage{latexsym}
\usepackage{lipsum}
\usepackage{hyperref}
\usepackage{verbatim}
\usepackage{graphicx}
\usepackage{makecell}
\usepackage{todonotes}
\usepackage{CJKutf8}

\usepackage{amssymb}
\usepackage{enumitem}
\usepackage{listings}
\usepackage{xstring}
\usepackage{graphicx}
\usepackage{pbox}
\usepackage{subcaption}
\usepackage{epstopdf}
\usepackage{xstring}

\usepackage{rotating}
\usepackage{xspace}
\usepackage{fontawesome}
\usepackage{amssymb, amsmath}
\usepackage{pbox}
\usepackage{booktabs} 
\usepackage{fancyvrb}
\usepackage{verbatimbox}
\usepackage{tablefootnote}
\usepackage{colortbl}

\usepackage{balance}
\usepackage{pbox}
\usepackage{multirow}
\usepackage{enumitem}
\usepackage[normalem]{ulem}
\usepackage{multicol}
\usepackage{tabularx}
\usepackage{tcolorbox}
\usepackage{color, colortbl}
\usepackage{soul}
\usepackage{booktabs}
\usepackage{multirow}
\usepackage[switch]{lineno}

\usepackage{comment}

\newcommand{\sq}{\faCheckSquare}
\newcommand{\cq}{\faCheck}

\usepackage[T1]{fontenc}

\usepackage[utf8]{inputenc}

\usepackage{microtype}

\usepackage{cancel}
\usepackage{ulem,xpatch}

\xpatchcmd{\sout}
  {\bgroup}
  {\bgroup}
  {}{}

\newcommand{\eat}[1]{} 

\title{Findings of the NLP4IF-2021 Shared Tasks\\ on Fighting the COVID-19 Infodemic and Censorship Detection}

\author{
Shaden Shaar,\textsuperscript{\rm 1} 
Firoj Alam,\textsuperscript{\rm 1} 
Giovanni Da San Martino,\textsuperscript{\rm 2}
Alex Nikolov,\textsuperscript{\rm 3}\\
{\bf Wajdi Zaghouani,\textsuperscript{\rm 4}
Preslav Nakov, \textsuperscript{\rm 1} \and
Anna Feldman\textsuperscript{\rm 5}  }\\
\textsuperscript{\rm 1} Qatar Computing Research Institute, HBKU, Qatar\\
\textsuperscript{\rm 2} University of Padova, Italy,\hspace{5mm}
\textsuperscript{\rm 3} Sofia University ``St. Kliment Ohridski'', Bulgaria,\\
\textsuperscript{\rm 4} Hamad bin Khalifa University, Qatar,\hspace{5mm}
\textsuperscript{\rm 5} Montclair State University, USA\\
\texttt{\{sshaar,fialam,wzaghouani,pnakov\}@hbku.edu.qa}\\ \texttt{dasan@math.unipd.it}\hspace{5mm}\texttt{feldmana@montclair.edu}\\
}

\begin{document}
\maketitle

\begin{abstract}
We present the results and the main findings of  the NLP4IF-2021 shared tasks. Task 1 focused on fighting the COVID-19 infodemic in social media, and it was offered in Arabic, Bulgarian, and English. Given a tweet, it asked to predict whether that tweet contains a verifiable claim, and if so, whether it is likely to be false, is of general interest, is likely to be harmful, and is worthy of manual fact-checking; also, whether it is harmful to society, and whether it requires the attention of policy makers.
Task~2 focused on censorship detection, and was offered in Chinese. A total of ten teams submitted systems for task 1, and one team participated in task 2; nine teams also submitted a system description paper.
Here, we present the tasks, analyze the results, and discuss the system submissions and the methods they used. Most submissions achieved sizable improvements over several baselines, and the best systems used pre-trained Transformers and ensembles. The data, the scorers and the leaderboards for the tasks are available at \url{http://gitlab.com/NLP4IF/nlp4if-2021}.
\end{abstract}

\section{Introduction} 
\label{sec:intro}

Social media have become a major communication channel, enabling fast dissemination and consumption of information. A lot of this information is true and shared in good intention; however, some is false and potentially harmful. While the so-called ``fake news'' is not a new phenomenon, e.g., the term was coined five years ago, the COVID-19 pandemic has given rise to the first global social media infodemic. The infodemic has elevated the problem to a whole new level, which goes beyond spreading fake news, rumors, and conspiracy theories, and extends to promoting fake cure, panic, racism, xenophobia, and mistrust in the authorities, among others. Identifying such false and potentially malicious information in tweets is important to journalists, fact-checkers, policy makers, government entities, social media platforms, and society. 

A number of initiatives have been launched to fight this infodemic, e.g.,~by building and analyzing large collections of tweets, their content, source, propagators, and spread~\citep{leng2020analysis,Medford2020.04.03.20052936,mourad2020critical,karami2021identifying}. Yet, these efforts typically focus on a specific aspect, rather than studying the problem from a holistic perspective. Here we aim to bridge this gap by introducing a task that asks to predict whether a tweet contains a verifiable claim, and if so, whether it is likely to be false, is of general interest, is likely to be harmful, and is worthy of manual fact-checking; also, whether it is harmful to society, and whether it requires the attention of policy makers. The task follows an annotation schema proposed in \cite{alam2020fighting,alam2021icwsmfighting}.

While the COVID-19 infodemic is characterized by insufficient attention paid to the problem, there are also examples of the opposite: tight control over information. In particular, freedom of expression in social media has been supercharged by a new and more effective form of digital authoritarianism. Political censorship exists in many countries, whose governments attempt to conceal or to manipulate information to make sure their citizens are unable to read or to express views that are contrary to those of people in power. One such example is Sina Weibo, a Chinese microblogging website with over 500 million monthly active users, which sets strict control over its content using a variety of strategies to target censorable posts, ranging from keyword list filtering to individual user monitoring: among all posts that are eventually censored, nearly 30\% are removed within 5–30 minutes, and for 90\% this is done within 24 hours \cite{zhu-etal:2013}. We hypothesize that the former is done automatically, while the latter involves human censors. Thus, we propose a shared task that aims to study the potential for automatic sensorship, which asks participating systems to predict whether a Sina Weibo post will be censored.

\section{Related Work}
\label{sec:related_work}
In this section, we discuss studies relevant to the COVID-19 infodemic and to censorship detection. 

\subsection{COVID-19 Infodemic}
Disinformation, misinformation, and ``fake news'' thrive in social media.
\citet{Lazer1094} and \citet{Vosoughi1146} in \emph{Science} provided a general discussion on the science of ``fake news'' and the process of proliferation of true and false news online.
There have also been several interesting surveys, e.g.,~\citet{Shu:2017:FND:3137597.3137600} studied how information is disseminated and consumed in social media. Another survey by \citet{thorne-vlachos:2018:C18-1} took a fact-checking perspective on ``fake news'' and related problems. Yet another survey \cite{Li:2016:STD:2897350.2897352} covered truth discovery in general. 
Some very recent surveys focused on stance for misinformation and disinformation detection \cite{Survey:2021:Stance:Disinformation}, on automatic fact-checking to assist human fact-checkers \cite{Survey:2021:AI:Fact-Checkers}, on predicting the factuality and the bias of entire news outlets \cite{Survey:2021:Media:Factuality:Bias}, on multimodal disinformation detection \cite{Survey:2021:Multimodal:Disinformation}, and on abusive language in social media \cite{Survey:2021:Abusive:Language}.

A number of Twitter datasets have been developed to address the COVID-19 infodemic. Some are without labels, other use distant supervision, and very few are manually annotated. \citet{cinelli2020covid19} studied COVID-19 rumor amplification in five social media platforms; their data was labeled using distant supervision. Other datasets include a multi-lingual dataset of 123M tweets~\cite{info:doi/10.2196/19273}, another one of 383M tweets \cite{Banda:2020}, a billion-scale dataset of 65 languages and 32M geo-tagged tweets~\cite{abdul2020mega}, and the GeoCoV19 dataset, consisting of 524M multilingual tweets, including 491M with GPS coordinates~\cite{Umair2020geocovid19}. There are also Arabic datasets, both with \cite{haouari2020arcov19:rumors,ARCorona:2021} and without manual annotations \cite{alqurashi2020large}.
We are not aware of Bulgarian datasets.

\citet{zhou2020repository} created the ReCOVery dataset, which combines 2,000 news articles about COVID-19, annotated for their factuality, with 140,820 tweets. \citet{vidgen2020detecting} studied COVID-19 prejudices using a manually labeled dataset of 20K tweets with the following labels: hostile, criticism, prejudice, and neutral.

\citet{song2020classification} collected a dataset of false and misleading claims about COVID-19 from IFCN Poynter, which they manually annotated with the following ten disinformation-related categories:
(1)~Public authority,
(2)~Community spread and impact, 
(3)~Medical advice, self-treatments, and virus effects,
(4)~Prominent actors,
(5)~Conspiracies,
(6)~Virus transmission,
(7)~Virus origins and properties,
(8)~Public reaction, and
(9)~Vaccines, medical treatments, and tests,
and
(10)~Cannot determine.

Another related dataset study by \cite{pulido2020covid} analyzed 1,000 tweets and categorized them based on factuality into the following categories: {\em (i)} False information, {\em (ii)} Science-based evidence, {\em (iii)} Fact-checking tweets, {\em (iv)} Mixed information, {\em (v)} Facts, {\em (vi)} Other, and {\em (vii)} Not valid. \citet{ding2020challenges} have a position paper discussing the challenges in combating the COVID-19 infodemic in terms of data, tools, and ethics. \citet{hossain2020covidlies} developed the COVIDLies dataset by matching a known misconceptions with tweets, and manually annotated the tweets with stance: whether the target tweet agrees, disagrees, or has no position with respect to a known misconception. Finally, \cite{Shuja2020.05.19.20107532} provided a comprehensive survey categorizing the COVID-19 literature into four groups: diagonisis related, transmission and mobility, social media analysis, and knowledge-based approaches.

The most relevant previous work is \cite{alam2021icwsmfighting,alam2020fighting}, where tweets about COVID-19 in Arabic and English were annotated based on an annotation schema of seven questions. Here, we adopt the same schema (but with binary labels only), but we have a larger dataset for Arabic and English, and we further add an additional language: Bulgarian.

\subsection{Censorship Detection}

There has been a lot of research aiming at developing strategies to detect and to evade censorship. Most work has focused on exploiting technological limitations with existing routing protocols \cite{leberknight-etal:2012a,katti-etal:2005,levin-etal:2015,weinberg-etal:2012,bock2020detecting}. Research that pays more attention to the linguistic properties of online censorship in the context of censorship evasion includes \citet{safaka-etal:2016}, who applied linguistic steganography to circumvent censorship. 

Other related work is that of \citet{lee:2016}, who used parodic satire to bypass censorship in China and claimed that this stylistic device delays and often evades censorship. \citet{hirun-etal:2015} showed that the use of homophones of censored keywords on Sina Weibo could help extend the time for which a Weibo post could remain available online. All these methods require significant human effort to interpret and to annotate texts to evaluate the likelihood of censorship, which might not be practical to carry out for common Internet users in real life.

\citet{king-etal:2013} in turn studied the relationship between political criticism and the chance of censorship. They came to the conclusion that posts that have a Collective Action Potential get deleted by the censors even if they support the state. \citet{zhang2019casm} introduced a system, Collective Action from Social Media (CASM), which uses convolutional neural networks on image data and recurrent neural networks with long short-term memory on text data in
a two-stage classifier to identify social media posts about offline collective action. \citet{zhang2019casm} found that despite online censorship in China suppressing the discussion of collective action in social media, censorship does not have a large impact on the number of collective action posts identified through CASM-China. \citet{zhang2019casm} claimed that the system would miss collective action taking place in ethnic minority regions, such as Tibet and Xinjiang, where social media penetration is lower and more stringent Internet control is in place, e.g.,~Internet blackouts.

Finally, there has been research that uses linguistic and content clues to detect censorship. \citet{knockel-etal:2015}  and \citet{zhu-etal:2013} proposed detection mechanisms to categorize censored content and to automatically learn keywords that get censored.
\citet{bamman-etal:2012} uncovered a set of politically sensitive keywords and found that the presence of some of them in a Weibo blogpost contributed to a higher chance of the post being censored. \citet{kei-nlp4if:2018} also targeted a set of topics that had been suggested to be sensitive, but unlike \citet{bamman-etal:2012}, they covered areas not limited to politics. \citet{kei-nlp4if:2018}, \citet{ng2019neural}, and \citet{ng2020linguistic} investigated how the textual content might be relevant to censorship decisions when both censored and uncensored blogposts include the same sensitive keyword(s).

\section{Tasks}
\label{sec:tasks}

Below, we describe the two tasks: their setup and their corresponding datasets.

\subsection{Task 1: COVID-19 Infodemic}
\label{ssec:covid19_infodemic}

\paragraph{Task Setup:}
The task asks to predict several binary properties for an input tweet about COVID-19. These properties are formulated in seven questions as briefly discussed below:
\begin{enumerate}[leftmargin=*]
\item \textbf{Verifiable Factual Claim:} \textit{Does the tweet contain a verifiable factual claim?}
A verifiable factual claim is a statement that something is true, and this can be verified using factual, verifiable information such as statistics, specific examples, or personal testimony. Following \cite{DBLP:journals/corr/abs-1809-08193}, factual claims could be ({\em a})~stating a definition, ({\em b})~mentioning a quantity in the present or in the past, ({\em c})~making a verifiable prediction about the future, ({\em d})~reference laws, procedures, and rules of operation, and ({\em e})~reference images or videos (e.g.,~``\emph{This is a video showing a hospital in Spain.}''), ({\em f})~implying correlation or causation (such correlation/causation needs to be explicit).

\item \textbf{False Information:} \textit{To what extent does the tweet appear to contain false information?}
This annotation determines how likely the tweet is to contain false information without fact-checking it, but looking at things like its style, metadata, and the credibility of the sources cited, etc.

\item  \textbf{Interesting for the General Public:} \textit{Will the tweet have an impact on or be of interest to the general public?}
In general, claims about topics such as healthcare, political news and findings, and current events are of higher interest to the general public. Not all claims should be fact-checked, for example ``\emph{The sky is blue.}'', albeit being a claim, is not interesting to the general public and thus should not be fact-checked.

\item \textbf{Harmfulness:} \textit{To what extent is the tweet harmful to the society/person(s)/company(s)/product(s)?}
The purpose of this question is to determine whether the content of the tweet aims to and can negatively affect the society as a whole, a specific person(s), a company(s), a product(s), or could spread rumors about them.\footnote{A rumor is a form of a statement whose veracity is not quickly or ever confirmed.}

\item \textbf{Need to Fact-Check:} \textit{Do you think that a professional fact-checker should verify the claim in the tweet?}
Not all factual claims are important or worth fact-checking by a professional fact-checker as this is a time-consuming process. For example, claims that could be fact-checked with a very simple search on the Internet probably do not need the attention of a professional fact-checker.

\item \textbf{Harmful to Society:} \textit{Is the tweet harmful for the society?} 
The purpose of this question is to judge whether the content of the tweet is could be potentially harmful for the society, e.g., by being weaponized to mislead a large number of people. For example, a tweet might not be harmful because it is a joke, or it might be harmful because it spreads panic, rumors or conspiracy theories, promotes bad cures, or is xenophobic, racist, or hateful.

\item \textbf{Requires Attention:} \textit{Do you think that this tweet should get the attention of government entities?} A variety of tweets might end up in this category, e.g., such blaming the authorities, calling for action, offering advice, discussing actions taken or possible cures, asking important questions (e.g.,~``\emph{Will COVID-19 disappear in the summer?}''), etc.
\end{enumerate}

\paragraph{Data:}
For this task, the dataset covers three different languages (Arabic, Bulgarian, and English), annotated with yes/no answers to the above questions. More details about the data collection and the annotation process, as well as statistics about the corpus can be found in \cite{alam2021icwsmfighting,alam2020fighting}, where an earlier (and much smaller) version of the corpus is described. We annotated additional tweets for Arabic and Bulgarian for the shared task using the same annotation schema. Table~\ref{tab:task1_distribution} shows the distribution of the examples in the training, development and test sets for the three languages. Note that, we have more data for Arabic and Bulgarian than for English.

\begin{table}[t]
\centering
\begin{tabular}{lrrrr}
\toprule
\multicolumn{1}{c}{\textbf{}} & \multicolumn{1}{c}{\textbf{Train}} & \multicolumn{1}{c}{\textbf{Dev}} & \multicolumn{1}{c}{\textbf{Test}} & \multicolumn{1}{c}{\textbf{Total}} \\ \midrule
Arabic & 520 & 2,536 & 1,000 & 4,056 \\
Bulgarian & 3,000 & 350 & 357 & 3,707 \\
English & 867 & 53 & 418 & 1,338 \\
\bottomrule
\end{tabular}
\caption{\textbf{Task 1}: Statistics about the dataset.}
\label{tab:task1_distribution}
\end{table}

\subsection{Task 2: Censorship Detection}
\label{ssec:censorship_detection}

\paragraph{Task Setup:}
For this task, we deal with a particular type of censorship -- when a post gets removed from a social media platform semi-automatically based on its content. The goal is to predict which posts on Sina Weibo, a Chinese microblogging platform, will get removed from the platform, and which posts will remain on the website.

\paragraph{Data:} Tracking censorship topics on Sina Weibo is a challenging task due to the transient nature of censored posts and the scarcity of censored data from well-known sources such as FreeWeibo\footnote{\url{http://freeweibo.com}} and WeiboScope\footnote{\url{http://weiboscope.jmsc.hku.hk}}. The most straightforward way to collect data from a social media platform is to make use of its API. However, Sina Weibo imposes various restrictions on the use of its API\footnote{http://open.weibo.com/wiki/API\begin{CJK*}{UTF8}{gbsn}文档\end{CJK*}/en} such as restricted access to certain endpoints and restricted number of posts returned per request. Above all, their API does not provide any endpoint that allows easy and efficient collection of the target data (posts that contain sensitive keywords). Therefore, \citet{ng2019neural} and \citet{ng2020linguistic} developed an alternative method to track censorship for our purposes. The reader is referred to the original articles to learn more details about the data collection. In a nutshell, the dataset contains censored and uncensored tweets, and it includes no images, no hyperlinks, no re-blogged content, and no duplicates.

For the present shared task 2, we use the balanced dataset described in \cite{ng2020linguistic} and \cite{ng2019neural}. The data is collected across ten topics for a period of four months: from August
29, 2018 till December 29, 2018. Table~\ref{tab:task2_distribution} summarizes the datasets in terms of number of censored and uncensored tweets in the training, development, and testing sets, while Table~\ref{tab:scraped-table} shows the main topics covered by the dataset.

\begin{table}[t]
\centering
\resizebox{200pt}{!}{
\begin{tabular}{lrrrr}
\toprule
\multicolumn{1}{c}{\textbf{}} & \multicolumn{1}{c}{\textbf{Train}} & \multicolumn{1}{c}{\textbf{Dev}} & \multicolumn{1}{c}{\textbf{Test}} & \multicolumn{1}{c}{\textbf{Total}} \\ \midrule
censored & 762 & 93 & 98& 953 \\
uncensored & 750 & 96 & 91 & 937 \\
Total   & 1,512 & 189 & 189 & 1,890 \\
\bottomrule
\end{tabular}
}
\caption{\textbf{Task 2}: Statistics about the dataset.}
\label{tab:task2_distribution}
\end{table}

\begin{table}[t] 
    \centering
     \resizebox{220pt}{!}{
    \begin{tabular}{lrr}
    \toprule
    \textbf{Topic} & \textbf{Censored} & \textbf{Uncensored}\\
    \midrule
        {cultural revolution} & 55 & 60  \\
        {human rights}& 53 & 67 \\
        {family planning}& 15 & 25  \\
        {censorship \& propaganda}& 32 & 54 \\
        {democracy}& 119 & 107 \\
        {patriotism}& 70 & 105 \\
        {China}& 186 & 194 \\
        {Trump}& 320 & 244 \\
        {Meng Wanzhou}& 55 & 76 \\
        {kindergarten abuse}& 48 & 5  \\
        \midrule
        {\textbf{Total}} & \textbf{953} & \textbf{937}\\
        \bottomrule
    \end{tabular}
    }
    \caption{\textbf{Task 2}: Topics featured in the dataset.}
    \label{tab:scraped-table}
\end{table} 

\section{Task Organization}
\label{sec:evaluation_framework}

In this section, we describe the overall task organization, phases, and evaluation measures.

\subsection{Task Phases}
\label{task_organization}
We ran the shared tasks in two phases:

\paragraph{Development Phase} In the first phase, only training and development data were made available, and no gold labels were provided for the latter. The participants competed against each other to achieve the best performance on the development set.

\paragraph{Test Phase} In the second phase, the test set (unlabeled input only) was released, and the participants were given a few days to submit their predictions.

\subsection{Evaluation Measures}
\label{ssec:evaluation_measures}

The official evaluation measure for task 1 was the average of the weighted F1 scores for each of the seven questions; for task 2, it was accuracy.

\section{Evaluation Results for Task 1}
\label{sec:results}

Below, we describe the baselines, the evaluation results, and the best systems for each language.

\subsection{Baselines}
The baselines for Task 1 are (\emph{i})~majority class, (\emph{ii})~ngram, and (\emph{iii})~random. The performance of these baselines on the official test set is shown in Tables \ref{tab:results_task_1_english}, \ref{tab:results_task_1_arabic}, and \ref{tab:results_task_1_bulgarian}. 

\subsection{Results and Best Systems}
The results on the official test set for English, Arabic, and Bulgarian are reported in Tables \ref{tab:results_task_1_english}, \ref{tab:results_task_1_arabic}, and \ref{tab:results_task_1_bulgarian}, respectively. 
We can see that most participants managed to beat all baselines by a margin.

Below, we give a brief summary of the best performing systems for each language. 

\paragraph{The English Winner:} Team \textbf{TOKOFOU}~\cite{NLP4IF-2021-TOKOFOUTeam} performed best for English. They gathered six BERT-based models pre-trained in relevant domains (e.g.,~Twitter and COVID-themed data) or fine-tuned on tasks, similar to the shared task's topic (e.g.,~hate speech and sarcasm detection). They fine-tuned each of these models on the task 1 training data, projecting a label from the sequence classification token for each of the seven questions in parallel. After model selection on the basis of development set F1 performance, they combined the models in a majority-class ensemble.

\paragraph{The Arabic Winner:} Team \textbf{R00} had the best performing system for Arabic. They used an ensemble of the follwoing fine-tuned Arabic transformers: AraBERT~\cite{antoun-etal-2020-arabert}, Asafaya-BERT~\cite{safaya-etal-2020-kuisail}, ARBERT. In addition, they also experimented with MARBERT~\cite{abdulmageed2020arbert}.

\paragraph{The Bulgarian Winner:} We did not receive a submission for the best performing team for Bulgarian. The second best team, \textbf{HunterSpeechLab}~\cite{NLP4IF-2021-HunterSpeechLabTeam},
explored the cross-lingual generalization ability of multitask models trained from scratch (logistic regression, transformer encoder) and pre-trained models (English BERT, and mBERT) for deception detection.

\begin{table*}[!tbh]
\centering
\resizebox{450pt}{!}{%
\setlength{\tabcolsep}{3pt}
\begin{tabular}{@{}clrrr|rrrrrrr@{}}
\toprule
\multicolumn{1}{c}{\textbf{Rank}} & \multicolumn{1}{c}{\textbf{Team}} & \multicolumn{1}{c}{\textbf{F1}} & \multicolumn{1}{c}{\textbf{P}} & \multicolumn{1}{c|}{\textbf{R}} & \multicolumn{1}{c}{\textbf{Q1}} & \multicolumn{1}{c}{\textbf{Q2}} & \multicolumn{1}{c}{\textbf{Q3}} & \multicolumn{1}{c}{\textbf{Q4}} & \multicolumn{1}{c}{\textbf{Q5}} & \multicolumn{1}{c}{\textbf{Q6}} & \multicolumn{1}{c}{\textbf{Q7}} \\ \midrule
1 & TOKOFOU & \bf \underline{0.897} & 0.907 & \bf 0.896  & 0.835 & 0.913 & 0.978 & \bf 0.873 & \bf 0.882 & 0.908 & \bf 0.889 \\
2 & dunder\_mifflin & 0.891 & 0.907 & 0.878  & 0.807 & 0.923 & 0.966 & 0.868 & 0.852 & \bf 0.940 & 0.884 \\
3 & NARNIA & 0.881 & 0.900 & 0.879  & 0.831 & 0.925 & 0.976 & 0.822 & 0.854 & 0.909 & 0.849 \\
4 & InfoMiner & 0.864 & 0.897 & 0.848  & 0.819 & 0.886 & 0.946 & 0.841 & 0.803 & 0.884 & 0.867 \\
5 & advex & 0.858 & 0.882 & 0.864  & 0.784 & \bf 0.927 & 0.987 & 0.858 & 0.703 & 0.878 & 0.866 \\
6 & LangResearchLabNC & 0.856 & \bf 0.909 & 0.827  & \bf 0.842 & 0.873 & 0.914 & 0.829 & 0.792 & 0.894 & 0.849 \\
 & \it majority\_baseline & 0.830 & 0.786 & 0.883  & 0.612 & \bf 0.927 & \bf 1.000 & 0.770 & 0.807 & 0.873 & 0.821 \\
 & \it ngram\_baseline & 0.828 & 0.819 & 0.868  & 0.647 & 0.904 & 0.992 & 0.761 & 0.800 & 0.873 & 0.821 \\
7 & HunterSpeechLab & 0.736 & 0.874 & 0.684  & 0.738 & 0.822 & 0.824 & 0.744 & 0.426 & 0.878 & 0.720 \\
8 & spotlight & 0.729 & 0.907 & 0.676  & 0.813 & 0.822 & 0.217 & 0.764 & 0.701 & 0.905 & 0.877 \\
 & \it random\_baseline & 0.496 & 0.797 & 0.389 & 0.552 & 0.480 & 0.457 & 0.473 & 0.423 & 0.563 & 0.526 \\ \bottomrule
\end{tabular}
}
\caption{\textbf{Task 1, English:} Evaluation results. For Q1 to Q7, the results are in terms of weighted F1 score.}
\label{tab:results_task_1_english}
\end{table*}

\begin{table*}[!tbh]
\centering
\resizebox{450pt}{!}{%
\setlength{\tabcolsep}{3pt}
\begin{tabular}{@{}clrrr|rrrrrrr@{}}
\toprule
\multicolumn{1}{c}{\textbf{Rank}} & \multicolumn{1}{c}{\textbf{Team}} & \multicolumn{1}{c}{\textbf{F1}} & \multicolumn{1}{c}{\textbf{P}} & \multicolumn{1}{c|}{\textbf{R}}  & \multicolumn{1}{c}{\textbf{Q1}} & \multicolumn{1}{c}{\textbf{Q2}} & \multicolumn{1}{c}{\textbf{Q3}} & \multicolumn{1}{c}{\textbf{Q4}} & \multicolumn{1}{c}{\textbf{Q5}} & \multicolumn{1}{c}{\textbf{Q6}} & \multicolumn{1}{c}{\textbf{Q7}} \\ \midrule
1 & R00 & \bf \underline{0.781} & \bf 0.842 & \bf 0.763 & 0.843 & 0.762 & 0.890 & 0.799 & \bf 0.596 & \bf 0.912 & 0.663 \\
$*$ & iCompass & 0.748 & 0.784 & 0.737 & 0.797 & 0.746 & 0.881 & 0.796 & 0.544 & 0.885 & 0.585 \\
2 & HunterSpeechLab & 0.741 & 0.804 & 0.700 & 0.797 & 0.729 & 0.878 & 0.731 & 0.500 & 0.861 & \bf 0.690 \\
3 & advex & 0.728 & 0.809 & 0.753 & 0.788 & \bf 0.821 & \bf 0.981 & \bf 0.859 & 0.573 & 0.866 & 0.205 \\
4 & InfoMiner & 0.707 & 0.837 & 0.639 & \bf 0.852 & 0.704 & 0.774 & 0.743 & 0.593 & 0.698 & 0.588 \\
 & \it ngram\_baseline & 0.697 & 0.741 & 0.716 & 0.410 & 0.762 & 0.950 & 0.767 & 0.553 & 0.856 & 0.579 \\
5 & DamascusTeam & 0.664 & 0.783 & 0.677 & 0.169 & 0.754 & 0.915 & 0.783 & 0.583 & 0.857 & 0.589 \\
 & \it majority\_baseline & 0.663 & 0.608 & 0.751 & 0.152 & 0.786 & \bf 0.981 & 0.814 & 0.475 & 0.857 & 0.579 \\
6 & spotlight & 0.661 & 0.805 & 0.632 & 0.843 & 0.703 & 0.792 & 0.647 & 0.194 & 0.828 & 0.620 \\
 & \it random\_baseline & 0.496 & 0.719 & 0.412 & 0.510 & 0.444 & 0.487 & 0.442 & 0.476 & 0.584 & 0.533 \\ \bottomrule
\end{tabular}
}
\caption{\textbf{Task 1, Arabic:}
Evaluation results. For Q1 to Q7, the results are in terms of weighted F1 score (The team iCompass submitted their system after the deadline, and thus we rank them with a $*$).}
\label{tab:results_task_1_arabic}
\end{table*}

\begin{table*}[!tbh]
\centering
\resizebox{450pt}{!}{%
\setlength{\tabcolsep}{3pt}
\begin{tabular}{@{}clrrr|rrrrrrr@{}}
\toprule
\multicolumn{1}{c}{\textbf{Rank}} & \multicolumn{1}{c}{\textbf{Team}} & \multicolumn{1}{c}{\textbf{F1}} & \multicolumn{1}{c}{\textbf{P}} & \multicolumn{1}{c|}{\textbf{R}} & \multicolumn{1}{c}{\textbf{Q1}} & \multicolumn{1}{c}{\textbf{Q2}} & \multicolumn{1}{c}{\textbf{Q3}} & \multicolumn{1}{c}{\textbf{Q4}} & \multicolumn{1}{c}{\textbf{Q5}} & \multicolumn{1}{c}{\textbf{Q6}} & \multicolumn{1}{c}{\textbf{Q7}} \\ \midrule
1 & advex & \bf \underline {0.837} & \bf 0.860 & \bf 0.861 & 0.887 & \bf 0.955 & 0.980 & 0.834 & \bf 0.819 & \bf 0.678 & \bf 0.706 \\
2 & HunterSpeechLab & 0.817 & 0.819 & 0.837 & \bf 0.937 & 0.943 & 0.968 & \bf 0.835 & 0.748 & 0.605 & 0.686 \\
 & \it majority\_baseline & 0.792 & 0.742 & 0.855 & 0.876 & 0.951 & \bf 0.986 & 0.822 & 0.672 & 0.606 & 0.630 \\
 & \it ngram\_baseline & 0.778 & 0.790 & 0.808 & 0.909 & 0.919 & 0.949 & 0.803 & 0.631 & 0.606 & 0.630 \\
3 & spotlight & 0.686 & 0.844 & 0.648 & 0.832 & 0.926 & 0.336 & 0.669 & 0.687 & 0.650 & 0.700 \\
4 & InfoMiner & 0.578 & 0.826 & 0.505 & 0.786 & 0.749 & 0.419 & 0.599 & 0.556 & 0.303 & 0.631 \\
 & \it random\_baseline & 0.496 & 0.768 & 0.400 & 0.594 & 0.502 & 0.470 & 0.480 & 0.399 & 0.498 & 0.528 \\ \bottomrule
\end{tabular}
}
\caption{\textbf{Task 1, Bulgarian:} Evaluation results. For Q1 to Q7 results are in terms of weighted F1 score.}
\label{tab:results_task_1_bulgarian}
\end{table*}

\subsection{Summary of All Systems}\label{sec:systemsummary}

\paragraph{DamascusTeam} \cite{NLP4IF-2021-DamascusTeam} used a two-step pipeline, where the first step involves a series of pre-processing procedures to transform Twitter jargon, including emojis and emoticons, into plain text. In the second step, a version of AraBERT is fine-tuned and used to classify the tweets. Their system was ranked 5th for Arabic.

\paragraph{Team dunder\_mifflin}~\cite{NLP4IF-2021-DunderMifflinTeam}
built a multi-output model using task-wise multi-head attention for inter-task information aggregation. This was built on top of the representations obtained from RoBERTa. To tackle the small size of the dataset, they used back-translation for data augmentation. Their loss function was weighted for each output, in accordance with the distribution of the labels for that output. They were the runners-up in the English subtask with a mean F1-score of 0.891 on the test set, without the use of any task-specific embeddings or ensembles. 

\paragraph{Team HunterSpeechLab}~\cite{NLP4IF-2021-HunterSpeechLabTeam} participated in all three languages. 
They explored the cross-lingual generalization ability of multitask models trained from scratch (logistic regression, transformers) and pre-trained models (English BERT, mBERT) for deception detection. They were 2nd for Arabic and Bulgarian.

\paragraph{Team iCompass}~\cite{NLP4IF-2021-iCompassTeam} had a late submission for Arabic, and would have ranked 2nd.
They used contextualized text representations from ARBERT, MARBERT, AraBERT, Arabic ALBERT and BERT-base-arabic, which they fine-tuned on the training data for task 1. They found that BERT-base-arabic performed best.

\paragraph{Team InfoMiner}~\cite{NLP4IF-2021-InfoMinerTeam}
participated in all three subtasks, and were ranked 4th on all three. They used pre-trained transformer models, specifically BERT-base-cased, RoBERTa-base, BERT-multilingual-cased, and AraBERT. They optimized these transformer models for each question separately and used undersampling to deal with the fact that the data is imbalanced.

\paragraph{Team NARNIA}~\cite{NLP4IF-2021-NARNIATeam} experimented with a number of Deep Learning models, including different word embeddings such as Glove and ELMo, among others. They found that the BERTweet model achieved the best overall F1-score of 0.881, securing them the third place on the English subtask.

\paragraph{Team R00}~\cite{NLP4IF-2021-R00Team}
had the best performing system for the Arabic subtask. They used an ensemble of neural networks combining a linear layer on top of one out of the following four pre-trained Arabic language models: AraBERT, Asafaya-BERT, ARBERT. In addition, they also experimented with MARBERT. 

\paragraph{Team TOKOFOU}~\cite{NLP4IF-2021-TOKOFOUTeam} 
participated in English only and theirs was the winning system for that language. They gathered six BERT-based models pre-trained in relevant domains (e.g.,~Twitter and COVID-themed data) or fine-tuned on tasks, similar to the shared task's topic (e.g.,~hate speech and sarcasm detection). They fine-tuned each of these models on the task 1 training data, projecting a label from the sequence classification token for each of the seven questions in parallel. After carrying out model selection on the basis of the F1 score on the development set, they combined the models in a majority-class ensemble in order to counteract the small size of the dataset and to ensure robustness.

\subsection{Summary of the Approaches}

Tables \ref{tab:overview_task1_english}, \ref{tab:overview_task1_arabic} and \ref{tab:overview_task1_bulgarian} offer a high-level comparison of the approaches taken by the participating systems for English, Arabic and Bulgarian, respectively (unfortunately, in these comparisons, we miss two systems, which did not submit a system description paper). We can see that across all languages, the participants have used transformer-based models, monolingual or multilingual. In terms of models, SVM and logistic regression were used. Some teams also used ensembles and data augmentation. 

\begin{table}[h!]
\centering
\setlength{\tabcolsep}{1.2pt}    
\small
\scalebox{0.80}{
\begin{tabular}{l|ll|ll|ll|lll}
\toprule
\multicolumn{1}{c}{\textbf{Ranks Team~}} & \multicolumn{2}{c}{\textbf{Trans.}} & \multicolumn{2}{c}{\textbf{Models}} & \multicolumn{2}{c}{\textbf{Repres.}} & \multicolumn{3}{c}{\textbf{Misc}} \\ \midrule
\textbf{~} & \rotatebox{90}{\textbf{BERT}} & \rotatebox{90}{\textbf{RoBERTa}} & \rotatebox{90}{\textbf{Logistic Regression}} & \rotatebox{90}{\textbf{SVM}} & \rotatebox{90}{\textbf{ELMo}} & \rotatebox{90}{\textbf{GloVe}} & \rotatebox{90}{\textbf{Ensemble}} & \rotatebox{90}{\textbf{Under/Over-Sampling}} & \rotatebox{90}{\textbf{Data Augmentation}} \\ \midrule
1. TOKOFOU & \sq & ~ & ~ & ~ & ~ & ~ & \sq & \multicolumn{1}{c}{~} & ~ \\
2. dunder\_mifflin & ~ & \sq & ~ & ~ & ~ & ~ & ~ & ~ & \sq \\
3. NARNIA & \sq & ~ & ~ & \cq & \cq & \cq & \cq & ~ & ~ \\
4. InfoMiner & \sq & \cq & ~ & ~ & ~ & ~ & ~ & \sq & ~ \\
7. HunterSpeechLab & \sq & ~ & \sq & ~ & ~ & ~ & ~ & ~ & ~ \\ \bottomrule
\end{tabular}
}

\setlength{\tabcolsep}{1.2pt}
\begin{tabular}{@{}rl@{}}
1 & \cite{NLP4IF-2021-TOKOFOUTeam} \\
2 & \cite{NLP4IF-2021-DunderMifflinTeam} \\
3 & \cite{NLP4IF-2021-NARNIATeam} \\
4 & \cite{NLP4IF-2021-InfoMinerTeam} \\
7 &  \cite{NLP4IF-2021-HunterSpeechLabTeam} 
\end{tabular}

\caption{\textbf{Task 1:} Overview of the approaches used by the participating systems for \textbf{English}. \sq$=$part of the official submission; \cq$=$considered in internal experiments; \emph{Trans.} is for Transformers; \emph{Repres.} is for Representations. References to system description papers are shown below the table.
}
\label{tab:overview_task1_english}
\end{table}

\begin{table}[h!]
\centering
\setlength{\tabcolsep}{1.2pt}    
\small
\scalebox{0.80}{
\begin{tabular}{l|llllll|l|ll}
\toprule
\multicolumn{1}{c}{\textbf{Ranks Team}} & \multicolumn{6}{c}{\textbf{Trans.}} & \multicolumn{1}{c}{\textbf{Models}} & \multicolumn{2}{c}{\textbf{Misc}} \\ \midrule
\textbf{~} & \begin{sideways}\textbf{BERT multilingual}\end{sideways} & \begin{sideways}\textbf{AraBERT}\end{sideways} & \begin{sideways}\textbf{Asafaya-BERT}\end{sideways} & \begin{sideways}\textbf{ARBERT}\end{sideways} & \begin{sideways}\textbf{ALBERT}\end{sideways} & \begin{sideways}\textbf{MARBERT}\end{sideways} & \begin{sideways}\textbf{Logistic Regression}\end{sideways} & \begin{sideways}\textbf{Ensemble}\end{sideways} & \begin{sideways}\textbf{Under/Over-Sampling}\end{sideways} \\ \midrule
1. R00 & ~ & \sq & \sq & \sq & ~ & \sq & ~ & \sq & ~ \\
* iCampass & ~ & \sq & ~ & \cq & \cq & \cq & ~ & ~ & ~ \\
2. HunterSpeechLab & \sq & ~ & ~ & ~ & ~ & ~ & \sq & ~ & ~ \\
4. InfoMiner & \cq & \sq & ~ & ~ & ~ & ~ & ~ & ~ & \sq \\
5. DamascusTeam & ~ & ~ & ~ & \sq & ~ & ~ & ~ & ~ & ~ \\ \bottomrule
\end{tabular}
}

\setlength{\tabcolsep}{1.2pt}
\begin{tabular}{@{}rl@{}}
1 & \cite{NLP4IF-2021-R00Team} \\
$*$ & \cite{NLP4IF-2021-iCompassTeam} \\
2 & \cite{NLP4IF-2021-HunterSpeechLabTeam} \\
4 & \cite{NLP4IF-2021-InfoMinerTeam} \\
5 &  \cite{NLP4IF-2021-DamascusTeam} 
\end{tabular}

\caption{\textbf{Task 1:} Overview of the approaches used by the participating systems for \textbf{Arabic}. 
}
\label{tab:overview_task1_arabic}
\end{table}

\begin{table}[h!]
\centering
\setlength{\tabcolsep}{1.2pt}    
\small
\scalebox{0.80}{
\begin{tabular}{l|l|l|l}
\toprule
\multicolumn{1}{c}{\textbf{Ranks Team~}} & \multicolumn{1}{c}{\textbf{Trans.}} & \multicolumn{1}{c}{\textbf{Models}} & \multicolumn{1}{c}{\textbf{Misc}} \\ \midrule
~ & \begin{sideways}\textbf{BERT multilingual}\end{sideways} & \begin{sideways}\textbf{Logistic Regression}\end{sideways} & \begin{sideways}\textbf{Under/Over-Sampling}\end{sideways} \\ \midrule
2. HunterSpeechLab & \sq & \sq & ~ \\
4. InfoMiner & \sq & ~ & \sq \\ \bottomrule
\end{tabular}
}

\setlength{\tabcolsep}{1.2pt}
\begin{tabular}{@{}rl@{}}
2 &  \cite{NLP4IF-2021-HunterSpeechLabTeam} \\
4 & \cite{NLP4IF-2021-InfoMinerTeam}
\end{tabular}
\caption{\textbf{Task 1:} Overview of the approaches used by the participating systems for \textbf{Bulgarian}. 
}
\label{tab:overview_task1_bulgarian}
\end{table}


\begin{table*}[!tbh]
    \centering
    \begin{tabular}{lcccc}
    \toprule
    \textbf{Team} & \textbf{P} &  \textbf{R} &  \textbf{F1} & \textbf{A}\\
    \midrule
        NITK$\_$NLP & \makecell{c: 0.69\\u: 0.61} &\makecell{c: 0.56\\u: 0.73}  & \makecell{c: 0.62\\u: 0.66} & 0.64 \\
        \midrule
       Baseline from \citep{ng2020linguistic}  & \makecell{c: 0.82\\u: 0.76} &\makecell{c: 0.79\\u: 0.79}  & \makecell{c: 0.80 \\u: 0.77 } & 0.80 \\ \midrule
       Majority baseline & & & & 0.50    \\
     Human baseline \citep{ng2020linguistic}   & & &   & 0.24    \\ 
     \bottomrule 
    \end{tabular}\caption{\label{tab:nitk}\textbf{Task 2:} the NITK$\_$NLP team's results. Here: \emph{c} is censored and \emph{u} is uncensored.}
\end{table*}

\section{Evaluation Results for Task 2}

Below, we report the results for the baselines and for the participating system.

\subsection{Baselines} 
For task 2, we have three baselines as shown in Table~\ref{tab:nitk}: a majority class baseline, as before, and two additional baselines described in \cite{ng2020linguistic}. The first additional baseline is a human baseline based on crowdsourcing. The second additional baseline is a multilayer perceptron (MLP) using linguistic features as well as such measuring the complexity of the text, e.g.,~in terms of its readability, ambiguity, and idiomaticity. These features are motivated by observations that censored texts are typically more negative, more idiomatic, contain more content words and more complex semantic categories. Moreover, censored tweets use more verbs, which indirectly points to the Collective Action Potential. In contrast, uncensored posts are generally more positive, and contain words related to leisure, reward, and money.

\subsection{Results}
Due to the unorthodox application, and perhaps to the sensitivity of the data, task 2 received only one submission: from team NITK$\_$NLP. The team used a pre-trained XLNet-based Chinese model by \citet{Cui_2020}, which they fine-tuned for 20 epochs, using the Adam optimizer. 
The evaluation results for that system are shown in Table~\ref{tab:nitk}. We can see that while the system outperformed both the human baseline and the majority class baseline by a large margin, it could not beat the MLP baseline. 
This suggests that capturing the linguistic fingerprints of censorship might indeed be important, and thus probably should be considered, e.g.,~in combination with deep contextualized representations from transformers \citep{ng2018detecting,ng2019neural,ng2020linguistic}.

\section{Conclusion and Future Work}
\label{sec:conclutions}

We have presented the NLP4IF-2021 shared tasks on fighting the COVID-19 infodemic in social media (offered in Arabic, Bulgarian, and English) and on censorship detection (offered in Chinese).

In future work, we plan to extend the dataset to cover more examples, e.g., from more recent periods when the attention has shifted from COVID-19 in general to vaccines. We further plan to develop similar datasets for other languages.

\section*{Ethical Considerations}

While our datasets do not contain personally identifiable information, creating systems for our tasks could face a ``dual-use dilemma,'' as they could be misused by malicious actors. Yet, we believe that the need for replicable and transparent research outweigh concerns about dual-use in our case.

\section*{Acknowledgments}

We would like to thank Akter Fatema, Al-Awthan Ahmed, Al-Dobashi Hussein, El Messelmani Jana, Fayoumi Sereen, Mohamed Esraa, Ragab Saleh, and Shurafa Chereen for helping with the Arabic data annotations.

This research is part of the Tanbih mega-project,
developed at the Qatar Computing Research Institute, HBKU, which aims to limit the impact of ``fake news,'' propaganda, and media bias by making users aware of what they are reading.

This material is also based upon work supported by the US National Science Foundation under Grants No. 1704113 and No. 1828199.

This publication was also partially made possible by the innovation grant No. 21 -- Misinformation and Social Networks Analysis in Qatar from Hamad Bin Khalifa University’s (HBKU) Innovation Center. The findings achieved herein are solely the responsibility of the authors.

\bibliographystyle{acl_natbib}
\bibliography{bib/censorship,bib/main,bib/custom}

\end{document}